\documentclass[letterpaper, 10 pt, conference]{ieeeconf}  
\IEEEoverridecommandlockouts                              
\usepackage{graphicx} 
\usepackage{times}       
\usepackage{amsmath}     
\usepackage{amssymb}     
\usepackage{graphicx}
\usepackage{pdfpages}
\usepackage{algorithm}
\usepackage[font=small]{caption}
\usepackage{subfig}
\usepackage[noend]{algpseudocode}
\usepackage[bookmarks=true]{hyperref}
\usepackage{subcaption}
\usepackage[backend=biber,style=ieee,mincitenames=1,maxnames=20,maxcitenames=1,natbib=true]{biblatex}
\usepackage{booktabs}
\usepackage{multirow}
\usepackage{balance}
\usepackage[font=small]{caption}

\def\eqref#1{Eq.~(\ref{#1})}

\newcommand\etal{\emph{et al.}}

\title{\LARGE \bf A Robust Placeability Metric for \\ Model-Free Unified Pick-and-Place Reasoning}

\author{Anonymous Authors
\vspace{-0.5cm}
}
\author{Benno Wingender$^*$ \and Nils Dengler$^*$ \and Rohit Menon \and Sicong Pan \and Maren Bennewitz
  \thanks{All authors are with the Humanoid Robots Lab, University of Bonn, Germany. 
  M. Bennewitz, R. Menon, S. Pan, and N. Dengler, are additionally with the Lamarr Institute for Machine Learning and Artificial Intelligence and the Center for Robotics, Bonn, Germany. 
  This work has partially been funded by the German Federal Ministry of Research, Technology and Space~(BMFTR) under the Robotics Institute Germany (RIG). Generative AI was used for linguistic refinement in the preparation of this work.}
  }

\begin{document}
\maketitle
\thispagestyle{empty} 
\pagestyle{empty}

\begin{abstract}
Reliable manipulation of previously unseen objects remains a fundamental challenge for autonomous robotic systems operating in unstructured environments.
In particular, robust pick-and-place planning directly from noisy and only partial real-world observations, where object surfaces are inherently incomplete due to occlusions (e.g., bottom faces on a tabletop), is difficult. 
As a result, many existing methods rely on strong object priors~\mbox{(e.g., CAD models)} or to assume placement on continuous, flat support surfaces such as planar tabletops, without explicitly accounting for edge proximity or inclined supports.
In this work, we introduce a robust probabilistic placeability metric that evaluates 6D~object placement poses from partial observations by jointly scoring object stability and graspability from raw point cloud geometry.
Using this metric, we generate diverse multi-orientation placement candidates and condition grasp scoring on these placements, enabling model-free unified pick-and-place reasoning.
Simulation and real-robot experiments on unseen objects and challenging support geometries confirm that our metric yields accurate stability predictions and consistently improves end-to-end pick-and-place success by producing stable, collision-free grasp–place pairs directly from partial point clouds. \\
\url{https://github.com/HumanoidsBonn/Placeability}
\end{abstract}

\section{Introduction}
\label{sec:intro}
The ability to pick objects and place them at desired locations is central to many robotic applications, including warehouse logistics, household assistance, and healthcare~\mbox{\cite{li2024comprehensive, zhang2022robotic}}.
In practice, achieving reliable pick-and-place execution in real-world environments requires perceiving an object’s geometry and pose, planning stable grasps~\cite{song2025overview}, identifying safe placements within target regions~\cite{ValidatingObjectPlacements_Harada_2014, ObjectPlacementPlanning_Haustein_2019}, and executing these actions in confined spaces under uncertainty~\cite{shi2025vmf}.
Consequently, rather than treating grasping and placing as independent problems, recent work has begun to reason jointly over both stages~\cite{PickAndPlace_Shanthi_2024, PlacementAware_Park_2024, Pick2place_He_2023}, which we refer to as \textit{unified pick-and-place reasoning}.
When reasoning about the target placement happens already during grasp selection, robots can choose grasps that enable both successful picking and placement, while avoiding collisions early and reducing the need for re-grasping caused by environmental constraints.

\begin{figure}[t]
  \centering
 \includegraphics[width=.95\columnwidth, trim=0 185 0 130, clip]{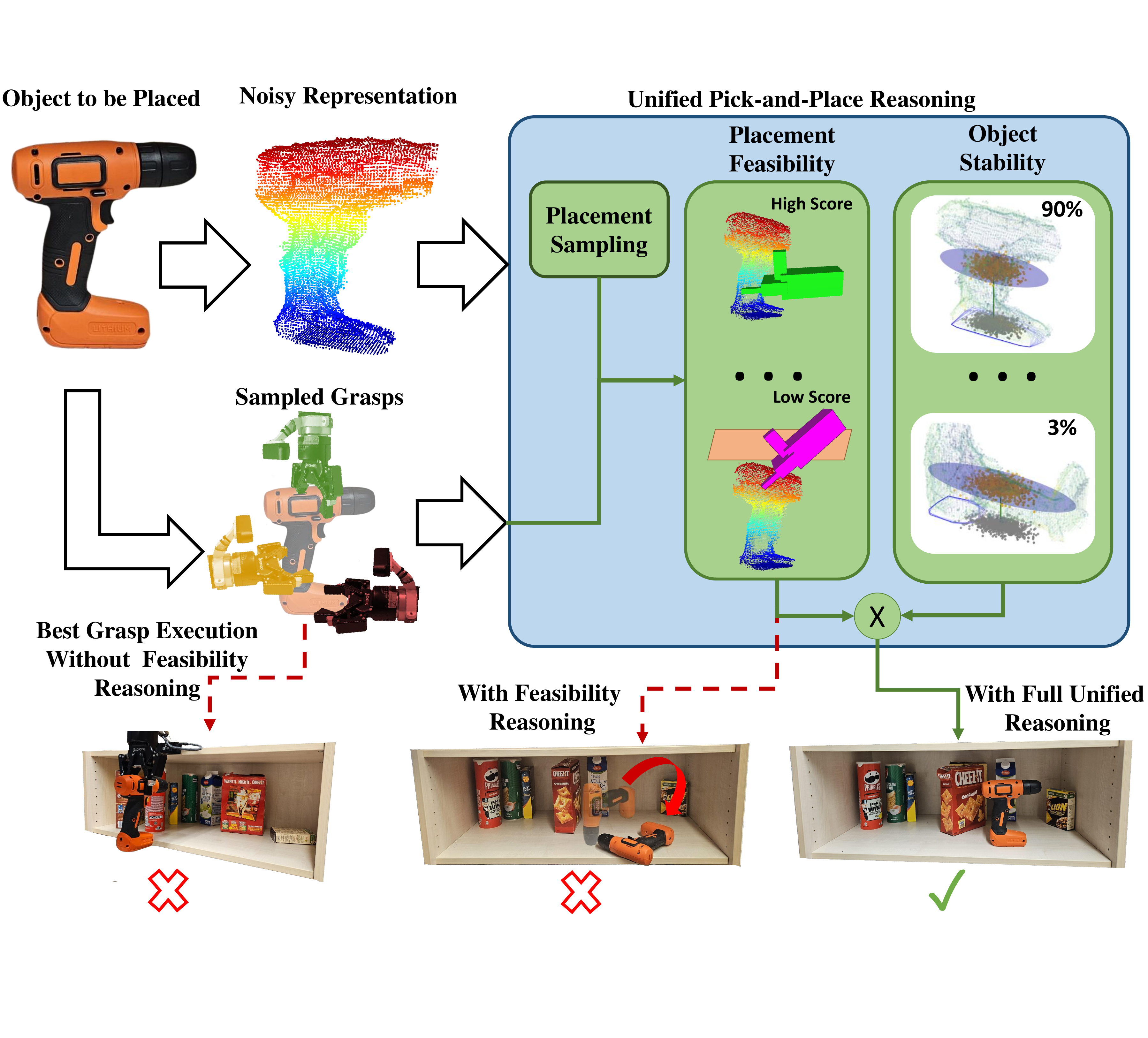}
\caption{
From a noisy point cloud reconstruction (top left), candidate grasps and placement poses are jointly evaluated using estimated stability and placement feasibility (top right). The system selects the highest-scoring grasp–placement pair that is executable and estimated to be physically stable at the target location (bottom right). Note that selecting the grasp based only on grasp quality leads to a collision with the shelf (bottom left) and choosing a feasible grasp without stability reasoning can result in object tipping~(bottom center). 
}
  \label{fig:cover}
  \vspace{-11px}
\end{figure}

However, existing work addressing object placement or joint pick-and-place planning largely falls into two groups.
First, \emph{model-free placement} methods can generalize to previously unseen objects from partial observations, but typically focus on predicting a stable placing surface or pose in isolation and do not directly support task-level pick-and-place reasoning under robot and environment constraints~\cite{noh2024learning, zhao2025anyplace}.
Second, \emph{unified pick-and-place} methods explicitly couple grasping and placing, but often assume complete object geometry (e.g., CAD models) and tabletop-style planar supports, while neglecting explicit placement stability or evaluating only a small set of predefined placements per grasp~\cite{ValidatingObjectPlacements_Harada_2014, ObjectPlacementPlanning_Haustein_2019, PickAndPlace_Shanthi_2024, PlacementAware_Park_2024, Pick2place_He_2023, lee2024spots}.
As a result, these approaches either lack a general, physically grounded stability score under realistic sensing noise or cannot robustly rank grasp–place pairs when object geometry and support conditions are only partially observed.
This gap is masked on tabletops but becomes a dominant failure mode in height-restricted shelves, where collision-free motion execution and object height severely restrict feasible grasp and placement pairs.
In such settings, the robot must reason jointly about where an object can be stably placed and which grasps remain executable at the target location under collisions and clearance constraints.

To address these limitations, we introduce a robust generalized \emph{placeability metric} that evaluates candidate 6D placement poses from partial point cloud observations by jointly scoring (i)~probabilistic stability under reconstruction uncertainty, and (ii)~placement-conditioned graspability (PCG) to ensure that grasps remain reachable and collision-free at the placement pose.
Using this metric, we generate diverse multi-orientation placement candidates from raw geometry and rank grasp-place pairs accordingly, enabling model-free unified pick-and-place reasoning for selecting stable and collision-free executions in constrained environments (Fig.~\ref{fig:cover}).

We evaluate our metric on unseen real objects, where object surfaces are inherently incomplete due to occlusions (e.g., bottom faces on a tabletop), and demonstrate improved performance over a CAD-based baseline~\cite{ObjectPlacementPlanning_Haustein_2019} in terms of stability prediction accuracy in challenging placement scenarios. 
Compared to a learning-based method~\cite{noh2024learning}, our approach achieves lower post-placement object pose error under partial and noisy observation conditions, as we show in a physics simulation.
Furthermore, we demonstrate the effectiveness of our placeability metric within a model-free unified pick-and-place pipeline through a  quantitative evaluation across varying scene configurations and object types.
Our contributions can be summarized as follows:
\begin{itemize}
\item A model-free placeability metric that evaluates 6D~object placement poses directly from partial point clouds by jointly reasoning about object stability and grasp feasibility for candidate poses, without requiring CAD models or predefined placements.
\item A probabilistic stability formulation that models both center-of-mass and placement surface contact uncertainty using TSDF weights as confidence, enabling physically grounded robust stability prediction under noisy observations of previously unseen objects.
\item A unified pick-and-place reasoning strategy that enables efficient selection of stable and executable grasp–place pairs in constrained environments, including cluttered and height-restricted shelves, where tabletop assumptions break.
\end{itemize}

\section{Related Work}
\label{sec:related}


\subsection{Placement Planning and Stability Evaluation}

Classical placement planning approaches rely on explicit geometric object models to evaluate placement stability~\mbox{\cite{ObjectPlacementInvMP_Holladay_2013, ValidatingObjectPlacements_Harada_2014, li2017visual, ObjectPlacementPlanning_Haustein_2019, nadeau2025generating, nadeau2025stable}}, or focus only on what side the object should be placed on ~\mbox{\cite{paxton2022predicting, noh2024learning, zhao2025anyplace}} instead of evaluating the overall object stability.
In early work, Harada~\etal~\cite{ValidatingObjectPlacements_Harada_2014} proposed to generate candidate placements by matching planar patches between object and environment polygon models and validating them via contact and stability tests. 
Similarly, Haustein~\etal~\cite{ObjectPlacementPlanning_Haustein_2019} presented a method to search for collision-free, reachable object poses with stable placement by projecting the center of mass onto convex hull faces while optimizing task-specific objectives such as object clearance. 
While physically grounded, these approaches assume complete object and environment geometry and are typically restricted to continuous support surfaces, neglecting edge placement scenarios.
More recently, generative and data-driven placement methods have been explored. 
Nadeau and Kelly~\cite{nadeau2025stable} proposed an inertia-aware placement planner that constructs robustness maps over support surfaces, while Nadeau~\etal~\cite{nadeau2025generating} extended this idea using diffusion models to generate placement pose candidates from point clouds. 
To reduce reliance on full object models, Noh~\etal~\cite{noh2024learning} introduced UOP-Net, that estimates stable placement regions directly from partial observations. 
Although effective, they assume that objects will always be placed on a planar table and do not provide a general placement quality metric across orientations or support geometries.
However, these methods still require object models for final placement validation.
Beyond geometry-driven placement, Zhao~\etal~\cite{zhao2025anyplace} developed AnyPlace, which combines vision–language models with diffusion-based pose prediction for context-driven placement. 
While demonstrating strong generalization across varying environments, the formulation is specialized to language-conditioned placement and does not explicitly model physical feasibility or robot constraints.
Lee \etal~\cite{lee2024spots} proposed SPOTS, a semi-autonomous framework that combines physics-based stability verification in simulation with LLM-based receptacle reasoning to recommend placement candidates. 
While their approach explicitly integrates physical robustness and contextual reasoning, it relies on CAD-based real-to-sim reconstruction and predefined object models for stability validation.

In contrast, we introduce a generalized, model-free placeability metric that evaluates placement feasibility directly from noisy point-cloud observations while accounting for estimated physical stability, clearance, and robot feasibility constraints.

\subsection{Unified Pick-and-Place Planning}
Pick-and-place planning can be formulated as sequential grasp-then-placement planning, or more recently as unified approaches that jointly reason about grasping and placement to improve task success and reduce replanning~\cite{PickAndPlace_Shanthi_2024, PlacementAware_Park_2024, Pick2place_He_2023, lobbezoo2021reinforcement, suomalainen2022survey, pantano2022capability, leebron2025b4p, maranci2024enabling}.
Most similar to our work, \mbox{Shanthi~\etal~\cite{PickAndPlace_Shanthi_2024}} formulated pick-and-place as a constrained optimization problem that maximizes joint grasp and placement success probability. 
While effective, the method relies on accurate learned success models and can be sensitive to prediction errors. 
He~\etal~\cite{Pick2place_He_2023} proposed an object-centric unified action space based on neural radiance fields~(NeRF), enabling joint reasoning over grasp and placement feasibility. 
However, the use of NeRF-based scene representations introduces high computational cost, limiting online applicability.
Qin~\etal~\cite{qin2025learning} presented an approach that predicts shared grasps that remain feasible at both pick and placement poses using energy-based models. 
While accelerating unified planning, the approach assumes known object meshes, fixed environments, and offline precomputed grasp candidates generated from CAD models.

In contrast, our approach enables unified pick-and-place reasoning directly from raw point cloud observations by evaluating placement feasibility first and then verifying grasp feasibility at candidate placement poses. 
This allows efficient evaluation of large placement candidate sets without requiring heavy prediction networks or object models, making the method well suited for online deployment in cluttered environments.

\begin{figure*}[t]
    \centering
    \includegraphics[width=\linewidth, trim= 0 15 95 0, clip]{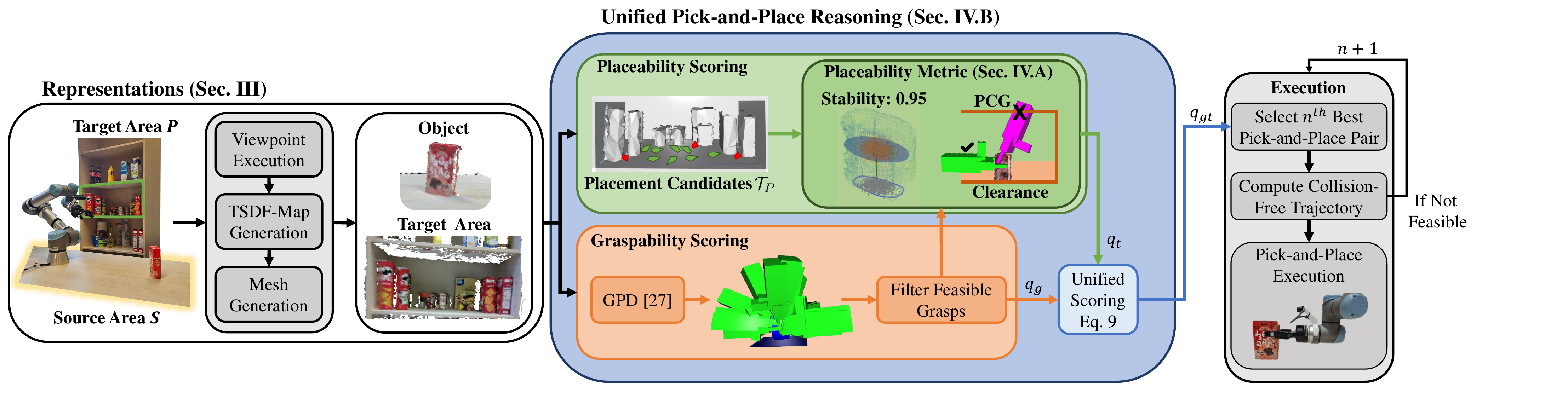}
    \caption{
    Overview of our framework for model-free unified pick-and-place reasoning.
    The pipeline integrates perception, grasp generation, and placement evaluation into a single object-centric process. 
    Point clouds are reconstructed from RGB-D observations to generate object and environment representations. 
    Feasible grasps and candidate placements are then evaluated using our generalized placeability metric, which accounts for stability, and placement-conditioned graspability (PCG). 
    A unified score combines these factors with heuristic constraints to select the best-scoring pick-and-place pair. 
    Finally, motion planning and execution proceed by iterating through candidate pairs in descending score order until a feasible trajectory is found.}
    \label{fig:UP4}
    \vspace{-7px}
\end{figure*}

\section{Problem Formulation and Assumptions}
We consider the following problem: 
Given source and target regions \(S\) and \(P\) reconstructed from sensor observations, and an object \(o\) detected within \(S\) with unknown 6D pose \(m_o \in \mathrm{SE}(3)\), the task is to compute the best physically stable, reliably graspable, and collision-free placement pose~$t_o$ out of a set of  placement candidates  $\mathcal{T}_P \in \mathrm{SE}(3)$ inside~\(P\) without a predefined position, orientation, and object model. 
The target area may exhibit cluttered objects and tight spatial constraints, which influence kinematic feasibility and collision-free placement. 
Note that we assume that the target object can be sufficiently observed.

We reconstruct the workspace and the object online from \mbox{RGB-D} observations by fusing depth data into a truncated signed distance function (TSDF) scene mesh $\mathcal{M}$~\cite{OverConfidence_Marques_2024}. 
In particular, from an initially unknown environment, we observe the scene from predefined view poses around an initial guess of the object's position and additional view poses that can sufficiently observe the key parts of the environment.
Note that no a priori knowledge about the object's shape or orientation, as well as the poses of objects present in \(P\) is assumed.
This representation supplies CAD-free, per-object geometry for placeability evaluation, and a reliable collision mesh for the environment for motion feasibility checks.

\section{Our Approach}

In this work, we introduce a generalized placeability metric that evaluates 6D object placement poses and grasps directly from observed point cloud data. 
By jointly assessing object stability and grasp feasibility at the placement pose, the metric enables model-free unified pick-and-place reasoning without relying on predefined object models or manually specified physical properties.
Fig.~\ref{fig:UP4} outlines our overall pipeline.
In the following, we first introduce the components of the placeability metric and then describe the unified grasp–place selection strategy.


\subsection{Components of the Placeability Metric}
\label{sec:placeability}
To ensure robust object placement, we introduce an object-centric \emph{placeability metric} that quantifies the physical feasibility of a candidate placement pose directly from the object’s sensed geometry.
The metric integrates three complementary components: 
(i) A \textbf{point cloud based probabilistic stability} measure that assigns higher scores to candidate placement poses whose projected center-of-mass hypotheses lie within the placing candidate's support polygon, ensuring robust equilibrium even under partial or noisy observations.
(ii) A \textbf{placement-conditioned graspability} measure that evaluates whether high-quality grasps remain kinematically feasible and collision-free after being transformed into a candidate placement pose, thereby linking grasp success directly to the following placement task.

\subsubsection{\textbf{Probabilistic Object Stability}}
To find stable 6D~placement poses for an object~$o$, including placements on inclined surfaces or near environmental edges, we estimate the probability that the object remains in static equilibrium when placed at a candidate pose~$t_o$ using a \mbox{Monte Carlo sampling} strategy.
In particular, let $o_t$ be the object transformed to its pose at $t_o$, we then model the apriori unknown Center of Mass~(CoM) as a Gaussian distribution~\mbox{$\mathcal{N}\!\left(\mu_{\mathrm{CoM}}, \Sigma_{\mathrm{CoM}}\right)$},
where the mean and covariance are computed from confidence-weighted points~$p_i$ of the object's geometry~$PC(o_t)$:
\begin{equation}
\mu_{\mathrm{CoM}} =
\frac{\sum_i w_i p_i}{\sum_i w_i}
\quad
\Sigma_{\mathrm{CoM}} =
\sum_i \left(\frac{w_i}{\sum_j w_j}\right)^2 \sigma_i^2
\end{equation}
Here, $\sigma_i^2 =\frac{1}{w_i + \epsilon}$ is the estimated per-point variance and inversely proportional to the confidence values \(w_i\) that are derived from TSDF weights, such that points with higher reconstruction confidence contribute more strongly to the CoM estimate.

To estimate the object’s support regions which it can be safely placed on, we model potential contact points probabilistically and derive support polygons $SP(o_t)$ from these sampled contacts.
First, we extract contact candidates~$\mathcal{C}(o_t)$ of the object at its pose $t_o$
\begin{equation}
\begin{aligned}
\mathcal{C}(o_t)
&=
\left\{
p_i \in PC(o_t)
\;\middle|\;
z(p_i) \le z_{\min} + \zeta
\right\}, \\
z_{\min}
&=
\min\limits_{p_j \in PC(o_t)} z(p_j),
\end{aligned}
\end{equation}
where $z(p_i)$ denotes the height of point $p_i$ and $\zeta$ is a small vertical tolerance.
We then again assign each candidate a positional variance~\(\sigma_i^2\) derived from TSDF weights and convert these uncertainties into contact sampling probabilities using a softmax model. 
For each Monte Carlo sample $n = 1, \dots, N$, we draw a subset $\mathcal{S}_n \subseteq \mathcal{C}(o_t)$ according to these probabilities and define a support polygon as the convex hull ($\mathrm{CH}$) of their projections onto the support plane via a function $ \Omega_{\mathrm{sp}}(\cdot)$:
\begin{equation}
\mathit{SP}_n(o_t)
=
\mathrm{CH}\!\left(
\left\{ \Omega_{\mathrm{sp}}(s) \mid s \in \mathcal{S}_n \right\}
\right).
\end{equation}
This procedure yields a distribution over support polygons $\{\mathit{SP}_n(o_t)\}_{n=1}^{N}$ that captures uncertainty.

\begin{figure}[t]
\centering
\includegraphics[width=\linewidth,trim={0 0 0 0},clip]{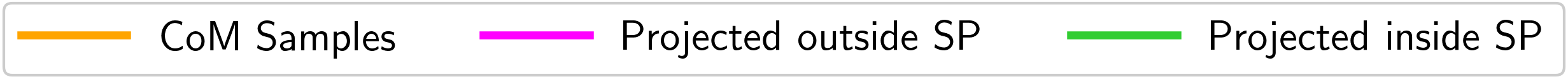}\\
\includegraphics[width=.32\linewidth,trim={180 20 200 70},clip]{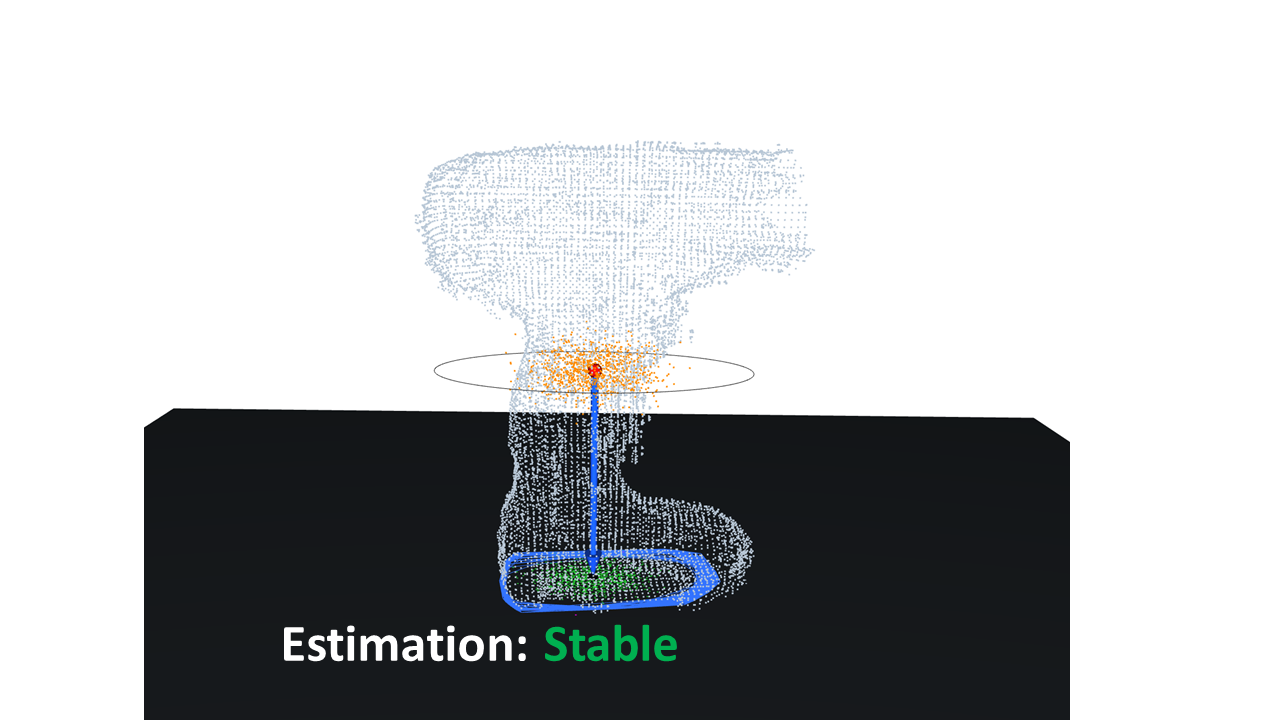}
\hfill
\includegraphics[width=.32\linewidth, trim={250 50 250 100},clip]{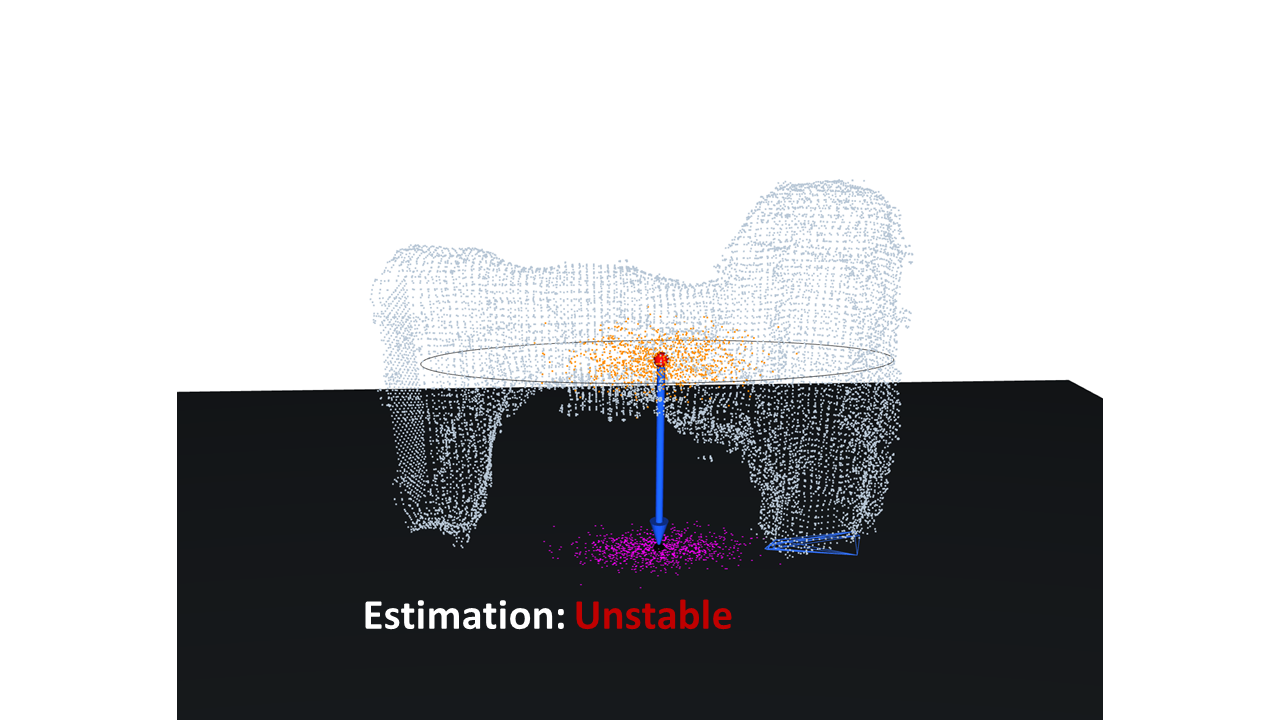}
\hfill
\includegraphics[width=.32\linewidth, trim={190 20 180 0},clip]{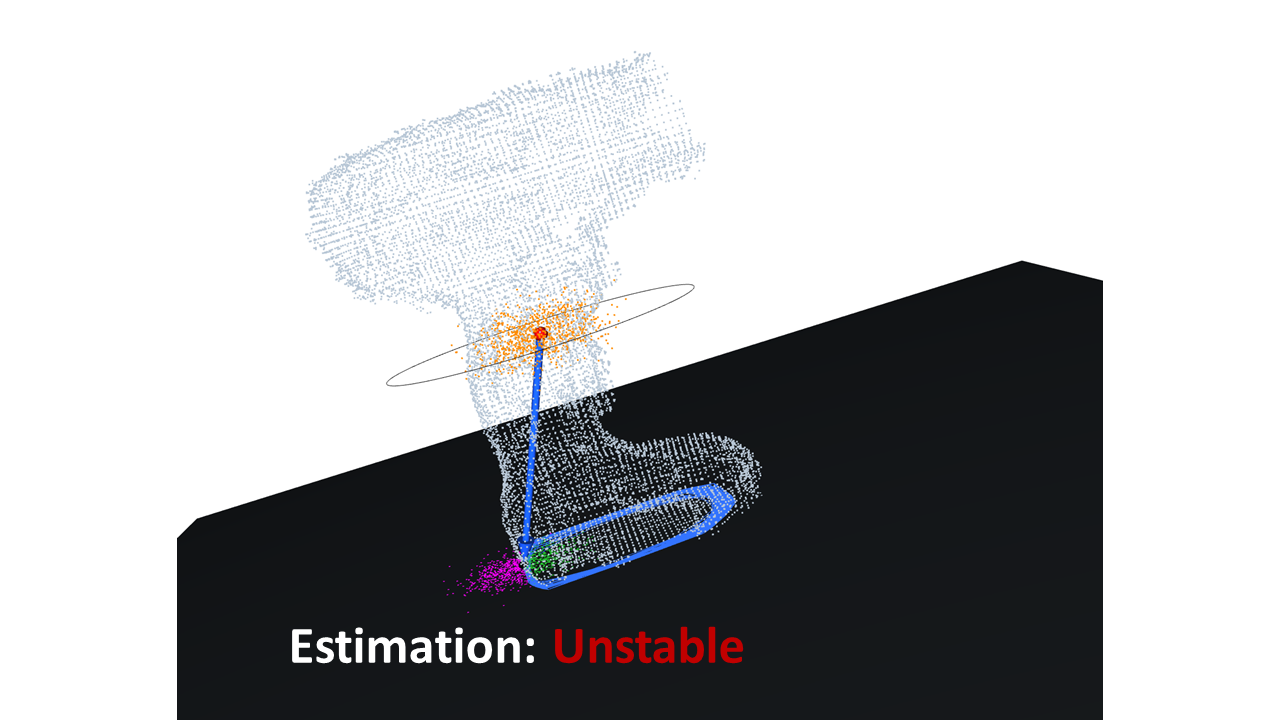}
\\
\includegraphics[width=.32\linewidth, trim={0 0 0 0},clip]{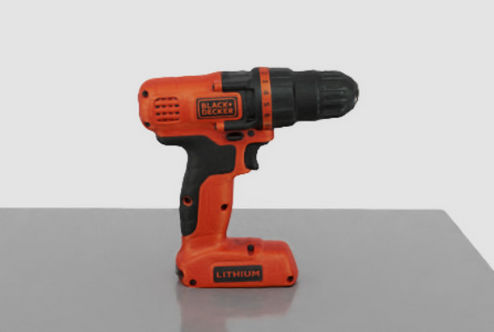}
\hfill
\includegraphics[width=.32\linewidth, trim={0 0 0 0},clip]{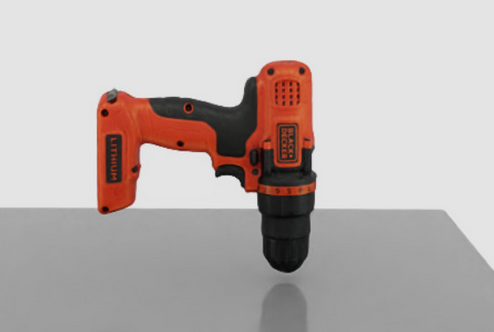}
\hfill
\includegraphics[width=.32\linewidth, trim={0 0 0 0},clip]{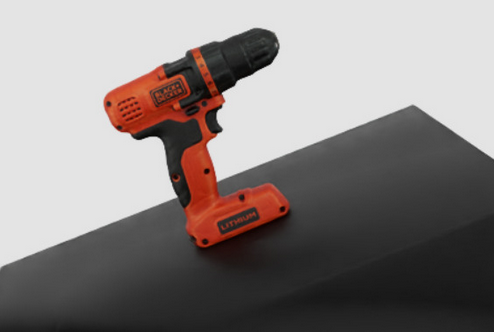} 

\caption{
Estimated stability illustrated on a noisy partial point cloud of a power drill  with its front section missing. \textit{Left:} most projected points lie inside the support polygon (stable). \textit{Center:} most projected points fall outside the support polygon (unstable). \textit{Right:} despite full contact, the steep tilt projects points beyond the polygon, resulting in predicted instability.
}
\label{fig:stability_example}
\vspace{-10px}
\end{figure}

Finally, we evaluate overall object stability by drawing, for each Monte Carlo sample $n$, a CoM candidate $c_n$ from its distribution together with a support polygon $\mathit{SP}_n(o_t)$ from the probabilistic contact model of the object.
We then project the CoM onto the support plane, and check whether $c_n$ lies inside the sampled support polygon.
Based on that, the stability probability is defined as:
\begin{equation}
f_{\mathrm{st}}(o_t)
=
\frac{1}{N}
\sum_{n=1}^{N}
\bigl(
\Omega_{\mathrm{sp}}(c_n)
\in
\mathit{SP}_n(o_t)
\bigr)
\end{equation}
The resulting stability value represents the probability that the object remains statically stable under reconstruction uncertainty and unknown internal mass distribution. 
Examples of our stability measure under different object orientations and environmental surface conditions are illustrated in Fig.~\ref{fig:stability_example}.

To improve robustness against execution errors and reconstruction noise of the target area, we additionally evaluate the stability under small random perturbations of the object pose. 
Specifically, we rotate the object point cloud $R$ times by randomly sampling pitch and roll offsets within $\pm 5^\circ$, corresponding to the maximum orientation deviation allowed by the motion planner, and compute the stability score for each perturbed pose.
We set $R=5$ empirically to balance robustness and computational cost.
The final probabilistic stability term is defined as the average over all trials:
\begin{equation}
\bar{f}_{\mathrm{st}}(o_t)
=
\frac{1}{R}
\sum_{r=1}^{R}
f_{\mathrm{st}}(o_t^{r})
\end{equation}

In summary, our formulation provides a probabilistic, uncertainty-aware stability estimate for arbitrary 6D~object placement  poses directly from reconstructed point clouds.
This enables robust stability reasoning without requiring complete object models or deterministic contact assumptions.

\subsubsection{\textbf{Placement-Conditioned Graspability (PCG)}}
To validate whether a candidate placement pose can actually be realized with a set of available grasp candidates, we introduce placement-conditioned graspability, which evaluates if a grasp remains feasible once transformed into the placing pose frame.
Let \(\mathcal{G}(o)=\{g_0, \dots, g_K\}\) be the set of $K$ candidate grasps for object $o$ at the object's observed pose~$m_o$.
Then, $T(t_o)$ denotes the rigid transform of a grasp~$g_k$ from its sampled grasp pose at~$m_o$ to a placement pose~$t_o$. 
The resulting transformed grasp, which we refer to as a~\mbox{\emph{place-grasp}}, is defined as $g_k^{t_o} = T(t_o)\,\cdot g_k$.

To compute the PCG, we score a grasp candidate~$g_k$ against a possible placement candidate $t_o \in \mathcal{T}_P$ to check if it remains reachable and collision-free:
\begin{equation}
\label{eq:pcg}
f_{\mathrm{pcg}}(g_k^{t_o})
=f_\mathrm{RM}(g_k^{t_o}) \cdot
f_\mathrm{coll}(g_k^{t_o}, \mathcal{M}) \cdot
f_{\mathrm{cl}}(g_k^{t_o})
\end{equation}
where \(f_\mathrm{RM}(\cdot)\in\{0,1\} \) tests kinematic feasibility for the manipulator via a precomputed reachability map~\cite{ReachabilityMap_Zacharias_2007} and \(f_\mathrm{coll}(\cdot,\mathcal{M})\in\{0,1\}\) checks a simplified gripper model against the environment mesh~ \(\mathcal{M}\) of the target region for collision. 
Lastly, to avoid collisions with the environmental supporting surface, we enforce a minimum vertical clearance ($\delta_{\min}$) between a candidate grasp (with vertical height $z(g_k^{t_o})$) and the object’s lowest sensed point~($z_{\min}$), with:
\begin{equation}
f_{\mathrm{cl}}(g_k^{t_o}) =
\begin{cases}
1, & \text{if } z(g_k^{t_o}) - z_{\min}. \ge \delta_{\min}, \\
0, & \text{otherwise}.
\end{cases}
\end{equation}
This formulation ensures that placement candidates are evaluated only based on grasps that remain kinematically feasible and collision-free at the target pose.

\subsection{Unified Pick-and-Place Reasoning}
\label{sec:unified_planning}

After introducing the components of our placeability metric, we now present our approach for model-free unified pick-and-place reasoning, which integrates grasp candidate selection and placement evaluation into a single pipeline. 

\subsubsection{\textbf{Placement Candidate Sampling}}
\label{sec:placement_sampling}
First, we compute the set of all possible placement candidates $\mathcal{T}_P$ from the simplified triangle mesh~$\mathcal{M}$ of $P$.
For that, we first extract horizontal surfaces of~$\mathcal{M}$ and sample a fixed set of points using Poisson-disk sample sets~\cite{yuksel2015sample}.
Furthermore, to increase orientation diversity among the placement poses, we apply a random $yaw$ rotation to each sampled placement. 
Additionally, we consider not just the initially observed orientation but also poses rotated around $\pm 90^\circ$ and $180^\circ  $ around the $pitch$ and $roll$ axes of the object, resulting in a total of six candidate orientations.
For each sampled pose, we then verify whether the object's support polygon at the corresponding pose lies within the convex boundary of the environment. 

\subsubsection{\textbf{Graspability Scoring}}
For the generation of grasp candidates on object point clouds, we utilize the Grasp Pose Detection (GPD) network~\cite{GPD_TenPas_2017}. 
For the object $o$, a set of candidate grasps $\mathcal{G}(o)$ is computed, from which the top-$k$ grasps~\mbox{$\{{g_0},\dots, g_k\}\in \mathcal{G}(o)$} are retained based on their grasp scores $q_g(g_k)$ predicted by~GPD.
Note that we adopt GPD for its robustness and availability, but any grasp prediction method or network can be used to generate the necessary grasp in our pipeline, provided that the grasp can be represented by an SE(3) rigid body transformation. 

\subsubsection{\textbf{Placeability Scoring}}
\label{Approach_place_planning}
Finally, we define placeability for each candidate place-grasp~$g_k$ of an object $o_t$ transformed to a candidate placement pose~$t_o$ by combining probabilistic stability and placement conditioned graspability.
The overall placeability score is then given by
%

\begin{equation}
\label{eq:placeability}
  q_t(g_k, \mathcal{T}_P) =
  \frac{1}{|\mathcal{T}_P|}
  \sum_{t_o \in \mathcal{T}_P}
  \bar{f}_{\mathrm{st}}(o^{t_o})
  \cdot
  f_{\mathrm{pcg}}(g_k^{t_o})
\end{equation}

To rank grasps according to both pick and place feasibility, we define a unified grasp score that considers jointly the original grasp quality~$q_g(g_k)$ as well as the associated stability and feasibility of the possible placements:
\begin{equation}
\label{eq:weighting}
q_{gt} (g_k, \mathcal{T}_P) = q_g(g_k) \cdot  q_t(g_k, \mathcal{T}_P)
\end{equation}
%
This formulation favors grasps that exhibit high intrinsic quality and are associated with stable, accessible placements. However, for action execution the final action is not selected by executing the single highest-scoring candidate. Instead, the retained grasp candidates $g_k$ are sorted in descending order according to the unified grasp-place score $q_{gt}$. Starting from the highest-ranked grasp, we evaluate the corresponding placement candidates and perform final feasibility checks. If no feasible placement can be found for the current grasp, the method proceeds to the next grasp in the ranked list. Once a feasible grasp-placement combination is found, we select the pose with the highest $q_t$ score. 

\begin{figure}[t]
  \centering
 \includegraphics[width=1.0\linewidth, trim= 125 40 70 260, clip]{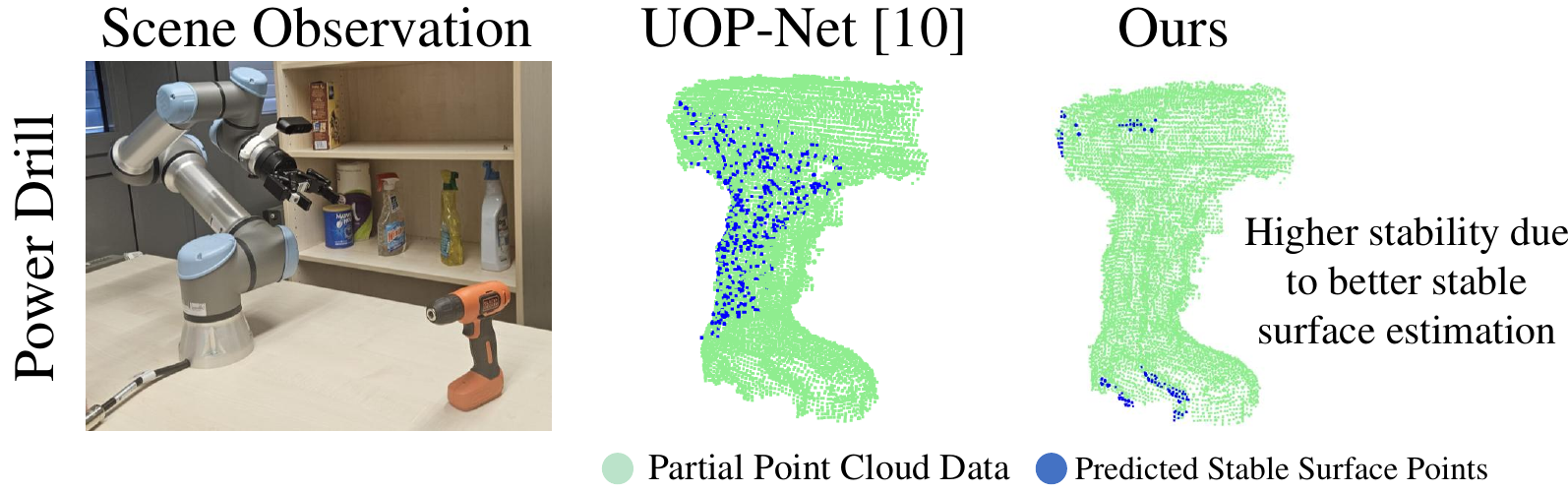}
\caption{
Qualitative comparison of predicted stable surface points from partial point cloud data by our method and \mbox{UOP-Net~\cite{noh2024learning}}.
As can be seen, the object cannot be placed on the plane derived from the surface points predicted by \mbox{UOP-Net} due to their curved geometry, whereas our method correctly identifies the peak contact points that allow for a stable placement. 
}
  \label{fig:vis_stable}
  \vspace{-10px}
\end{figure}
\section{Experimental Evaluation}
\label{sec:exp}

To validate the effectiveness of our approach, we evaluate the proposed placeability metric with respect to its core contributions: probabilistic stability estimation from partial point clouds, object-centric model-free reasoning, and unified pick-and-place reasoning.
All experiments were conducted using a UR5e 6-DoF arm with a Robotiq 2F-85 gripper and common household objects, spanning diverse geometries and mass distributions.
Perception relied on a wrist-mounted Orbbec Gemini 336 RGB-D camera and an external RealSense D435 camera.
The system ran on ROS~2 Humble on a workstation equipped with an NVIDIA RTX~4080~Super GPU, an AMD Ryzen~9~7900X3D CPU, and 128\,GB of RAM.
For the source area, we captured real-world partial point clouds from three fixed viewpoints and complemented them with observations from the external camera to extend the observable workspace. For stability estimation, we used $N=100$ Monte Carlo samples and set $\zeta=0.02\,\mathrm{m}$. For the feasibility checks, we used a vertical tolerance of \mbox{$\delta_{\min}=0.02\,\mathrm{m}$}.

\begin{table}[t]
\vspace{-3px}
\centering
\resizebox{\linewidth}{!}{%
\begin{tabular}{c|c|ccc}
\toprule
\textbf{Partial Point Cloud} & \textbf{Method} & \textbf{Rot. [deg] $\downarrow$} & \textbf{Trans. [cm] $\downarrow$} & \textbf{L2 Full $\downarrow$}\\
\midrule

\textbf{Pringles Can} & Ours     & $\bf8.56 \pm 12.12$ & $\bf0.27 \pm 0.23$ & $\bf0.21 \pm 0.30$ \\
Coverage: $84.14\% \pm 1.70$ & UOP-Net     & $8.88 \pm 7.48$ & $1.99 \pm 2.05$ & $0.22 \pm 0.18$ \\
\midrule

\textbf{Power Drill}
& Ours     & $\bf2.50 \pm 1.48$ & $\bf0.16 \pm 0.08$ & $\bf0.06 \pm 0.04$ \\
Coverage: $73.90\% \pm 1.94$& UOP-Net      & $8.15 \pm 3.10$ & $1.47 \pm 0.88$ & $0.20 \pm 0.08$ \\
\midrule

\textbf{Mustard Bottle}
& Ours      & $\bf6.81 \pm 2.82$ & $\bf0.37 \pm 0.27$ & $\bf0.17 \pm 0.07$ \\
Coverage: $84.86\%\pm 0.64$ & UOP-Net       & $8.02 \pm 6.22$ & $0.36 \pm 0.17$ & $0.20 \pm 0.15$ \\
\midrule

\textbf{Cracker Box}
& Ours     & $\bf2.80 \pm 0.79$ & $0.68 \pm 0.41$ & $\bf0.07 \pm 0.02$ \\
Coverage: $73.24\% \pm 6.12$ & UOP-Net       & $2.93 \pm 1.50$ & $\bf0.17 \pm 0.11$ & $0.07 \pm 0.04$ \\
\midrule
\midrule

\multirow{2}{*}{\textbf{Overall}}
& Ours       & $\bf5.16 \pm 6.34$ & $\bf0.37 \pm 0.32$ & $\bf0.13 \pm 0.15$ \\
& UOP-Net      & $6.99 \pm 5.32$ & $1.00 \pm 1.29$ & $0.17 \pm 0.13$ \\

\bottomrule
\end{tabular}}
\caption{
Placement stability evaluation on four YCB objects reconstructed from partial point clouds, with average ground-truth surface coverage reported per object. 
We compare our probabilistic placeability metric with UOP-Net~\cite{noh2024learning}, where lower rotational and translational pose deviations indicate more stable placements after physics simulation. 
Our method achieves the best overall performance and consistently reduces post-placement pose error compared to UOP-Net predictions, demonstrating robust stability estimation under partial and noisy observations.
}
\label{tab:pose_comparison_extended}
\vspace{-12px}
\end{table}

\begin{figure*}[t]
\centering
\vspace{-3px}

\begin{minipage}{0.65\textwidth}
\centering
\includegraphics[width=\linewidth]{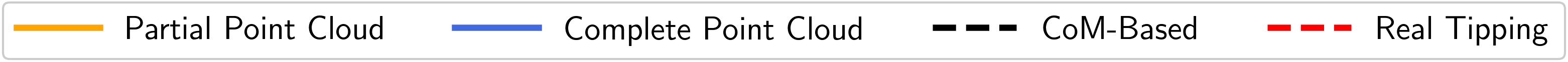}
\end{minipage}
\vspace{-10px}

\subfloat[Power Drill: Backwards\label{fig:stability:a}]
{\includegraphics[width=.25\textwidth]{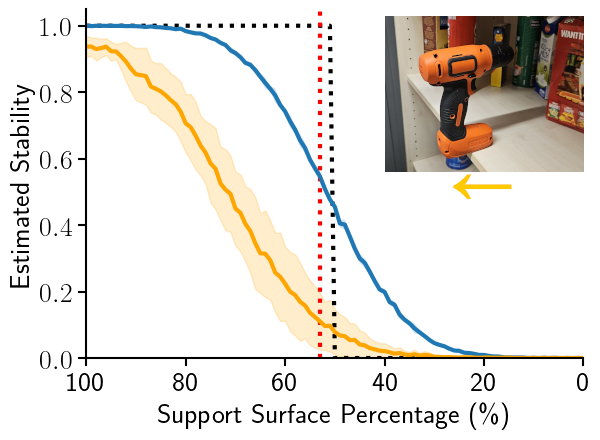}}
\subfloat[Power Drill: Forwards\label{fig:stability:b}]
{\includegraphics[width=.25\textwidth]{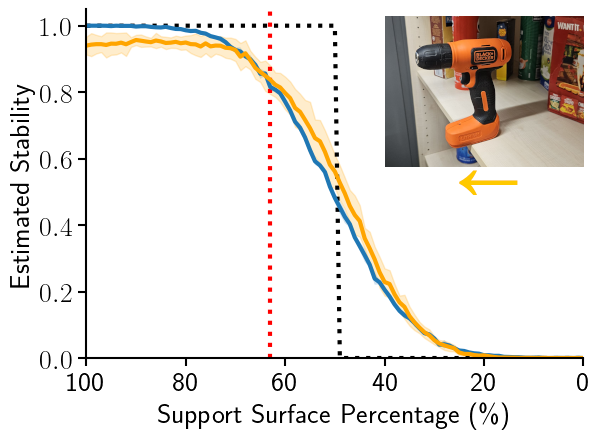}}
\subfloat[Cheez-It Box: Lying\label{fig:stability:c}]
{\includegraphics[width=.25\textwidth]{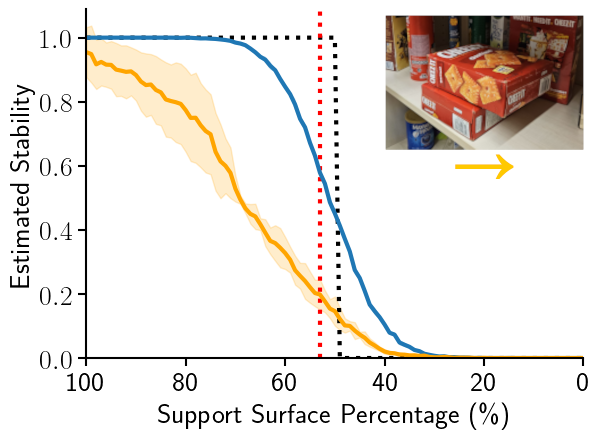}}
\subfloat[Surface Inclination\label{fig:stability:d}]
{
    \includegraphics[
        width=.25\textwidth,
        trim={250px 105px 250px 100px},
        clip
    ]{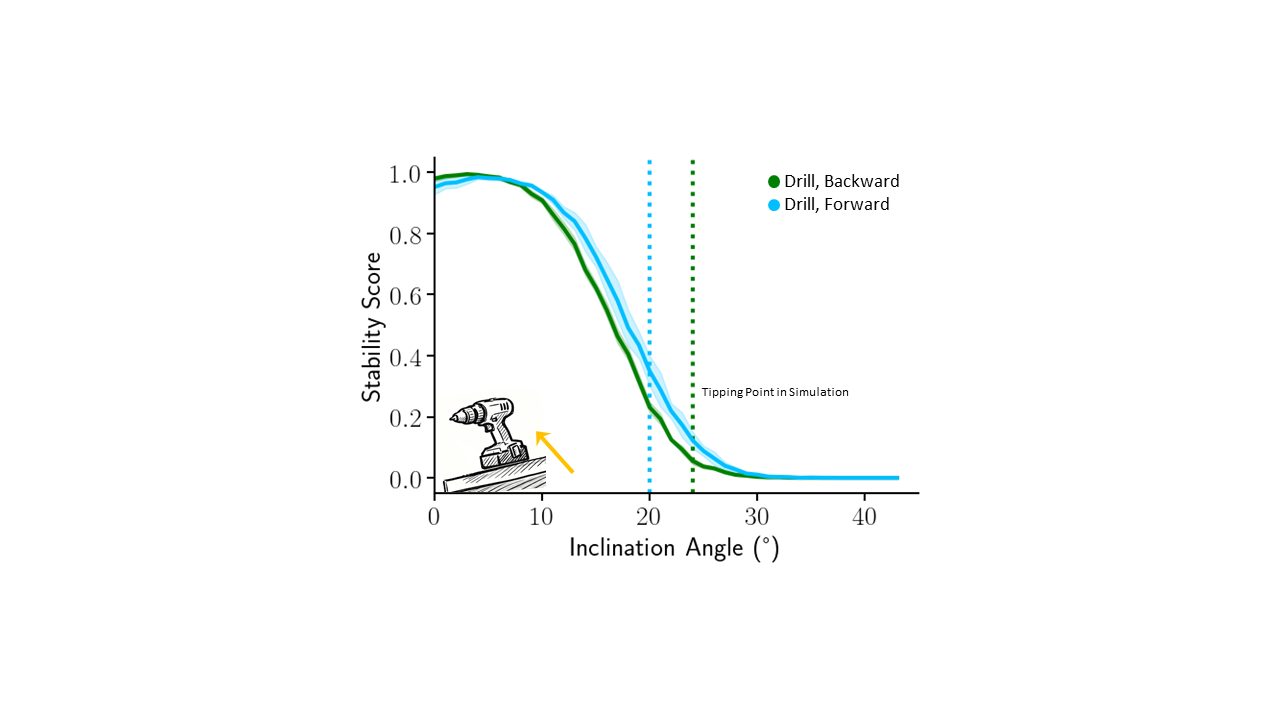}
}
\caption{
Edge-proximity and inclination tipping analysis on a real power drill (two orientations) and a real Cheez-It box. 
The plots show the predicted probabilistic stability score while translating the object toward a support surface edge~(a)--(c) and increasing the environmental support plane inclination~(d). 
Blue curves indicate evaluation using the complete ground-truth mesh, while all remaining curves use partial noisy point clouds reconstructed from sensor observations. 
Despite incomplete geometry, the predicted stability follows the ground-truth trend and transitions near the experimentally measured tipping thresholds. 
Deviations arise from missing contact regions and uncertainty in the reconstructed support polygon. 
The consistent drop of the stability score to approximately $0.5$ near the true tipping point indicates that our probabilistic formulation remains robust under realistic perception conditions without requiring CAD models.
}
\label{fig:stability}
\vspace{-8px}
\end{figure*}

\subsection{Quantitative Evaluation of Object Stability}
To assess the physical validity of our placeability metric, we first  evaluate the probabilistic stability estimate~$f_{\mathrm{st}}(o_t)$. 
We conduct two focused experiments to validate our stability formulation.

\subsubsection{\textbf{General Object Stability Prediction}}
We compare our probabilistic stability measure against the placing-surface prediction network UOP-Net~\cite{noh2024learning} on previously unseen objects, using the authors’ publicly available codebase and pretrained weights.
UOP-Net is the most closely related method in terms of stable placement prediction, because it operates on partial point clouds rather than CAD models and predicts a stable plane that approximates the object's support surface for placement on a horizontal tabletop.
We use four partial YCB~\cite{calli2015ycb} object point clouds, captured from four fixed viewpoints of our robotic system. 
To account for reconstruction inaccuracies and sensor noise, each object is recorded five times. 
This setup reflects realistic partial observations with different mesh coverages across objects, as summarized in Table~\ref{tab:pose_comparison_extended}.

For each object, we first extract the most stable surface points predicted by UOP-Net and our method. 
Afterwards, we fit a plane to these points to generate the predicted object surface the object should be placed on. 
Both predictions are evaluated in physics simulation by placing the object with the respective support surface in contact with a planar table and measuring whether it remains at rest. 
Stability is quantified using the positional offset metrics proposed in~\cite{noh2024learning}, computed between the placed object pose before and after simulation.
We perform this evaluation in simulation to accurately measure the resulting pose offset after placement. 

The results in Tab.~\ref{tab:pose_comparison_extended} show that our method outperforms UOP-Net on objects of complex geometry, while performance remains comparable and occasionally favors \mbox{UOP-Net} for simpler shapes.
Furthermore, we showcase a qualitative example of the predicted surface in Fig.~\ref{fig:vis_stable}, where our method results in a better stable surface prediction due to its curved geometry, placing the drill on the side with the outer points, while UOP would push the object slightly into the table.
Overall, these results indicate that our stability prediction is robust to partial and noisy observations and consistently provides more reliable placement estimates than UOP-Net, particularly for geometrically complex objects such as the drill where stable support regions cannot be approximated by a single planar surface.

\subsubsection{\textbf{Tipping Point Prediction}}
Beyond planar placements, a stability metric must reliably predict whether an object remains in static equilibrium on arbitrary supporting surfaces. 
We therefore evaluate our stability formulation in two challenging settings: (i) edge proximity, where we estimate the tipping threshold of an object approaching a support surface boundary 
and (ii) surface inclination, where we measure stability under increasing support tilt.

We evaluate the metric on a power drill and \mbox{Cheez-It box} and consider two orientations for the drill, as its center of mass is offset from the object’s geometric center, resulting in asymmetric tipping behavior. 
As a baseline, we compute tipping thresholds using the CAD model’s center of mass, following prior work that approximates the CoM from object geometry alone~\cite{ObjectPlacementPlanning_Haustein_2019}.
Note that UOP-Net~\cite{noh2024learning} cannot be evaluated in this setting, since it only predicts stable surface points for placement on a flat, tabletop-like support.
For edge proximity, we obtained real-world tipping thresholds by gradually translating each object toward a support boundary until static equilibrium was lost, and report the threshold as the percentage of the object’s support footprint remaining in contact with the surface.
As shown in Fig.~\ref{fig:stability}.a–c, our stability estimates exhibit a transition near the measured real-world tipping thresholds across all objects and orientations and correctly predict the earlier tipping of the power drill in the backward configuration. 
In contrast, the CAD-based CoM baseline (black dotted) deviates near the threshold, as it ignores contact geometry and partial support effects.

For object potential object stacking cases, we evaluate stability under increasing support tilt in simulation to obtain repeatable tipping angles. 
In particular, we initialize the object in a stable pose on a planar support and increment the support inclination in \mbox{$1^\circ$ steps} until static equilibrium is lost. 
As shown in Fig.~\ref{fig:stability}.d, the predicted stability decreases monotonically with increasing tilt and drops near the simulated tipping angle for the drills~($\approx 20^\circ$ and $23^\circ$). 
As our stability formulation does not model friction, it may slightly overestimate stability for objects that slip rather than topple, particularly when their support area is elongated downslope.
Nevertheless, these results demonstrate that our proposed metric reliably captures geometry- and center-of-mass–driven tipping behavior.

\begin{table*}[t]
\centering
\begin{tabular}{l|cccc||cccc}
\hline
 & \multicolumn{4}{c||}{\textbf{Cluttered Shelf Environment}} 
 & \multicolumn{4}{c}{\textbf{Height-Reduced Environment}} \\
\textbf{Object} 
 & Grasp-RP & Grasp-MO & UniP-NoStab & \textbf{UniP (Ours)}
 & Grasp-RP & Grasp-MO & UniP-NoStab & \textbf{UniP (Ours)}  \\
\hline
Drill    & 0\%   & 0\%   & 67\% & \textbf{100\%} & 33\% & 0\% & 33\% & \textbf{100\%}\\
Cereals  & 33\%  & 100\% & 100\% & \textbf{100\%} & 0\%  & 0\% & 100\% & \textbf{100\%}\\
Pringles & 67\%  & 100\% & 100\% & \textbf{100\%} & 0\%  & 0\% & 67\% & \textbf{67\%}\\
Mustard  & 100\% & 67\%  & 67\%  & \textbf{100\%} & 67\% & 100\% & 33\% & \textbf{67\%}\\
Milk     & 33\%  & 33\%  & 100\% & \textbf{67\%}  & 33\% & 0\% & 67\% & \textbf{100\%}\\
\hline
\textbf{Average} 
 & 46.6\% & 60.0\% & 86.8\% & \textbf{93.4\%}
 & 26.6\% & 20\% & 60.0\% & \textbf{86.8\%}\\
\hline
\textbf{Failure Type} 
 & \multicolumn{8}{c}{\textbf{Failure Breakdown}} \\
\hline
Grasp failed, place found
 & 2 & -- & -- & 1
 & 2 & 3 & 2 & 2 \\

Grasp valid, No place found
 & 6 & 4 & -- & --
 & 9 & 8 & -- & -- \\

Unstable placing 
 & -- & 1 & 2 & --
 & -- & 1 & 4 & -- \\
\hline
\end{tabular}
\caption{Real-world pick-and-place success rates across five objects evaluated and three executions per object in two shelf configurations. 
Our unified pipeline (UniP) achieves the highest overall performance in both scenarios, maintaining high success even under reduced vertical clearance where the sequential pick-then-place baselines Grasp-RP and Grasp-MO degrade substantially.
The failure breakdown shows that sequential approaches frequently fail after successful grasps due to missing feasible placements, while the unified ablation (UniP-NoStab) exhibits additional unstable placements but still outperforms the sequential baselines. 
These results demonstrate that combining unified grasp–placement reasoning with probabilistic stability estimation is critical for succesful manipulation in constrained environments.}
\label{tab:multi_env_eval_extended}
\vspace{-8px}
\end{table*}

\begin{figure}[t]
  \centering
 \includegraphics[width=1.0\linewidth, trim= 260 210 240 210, clip]{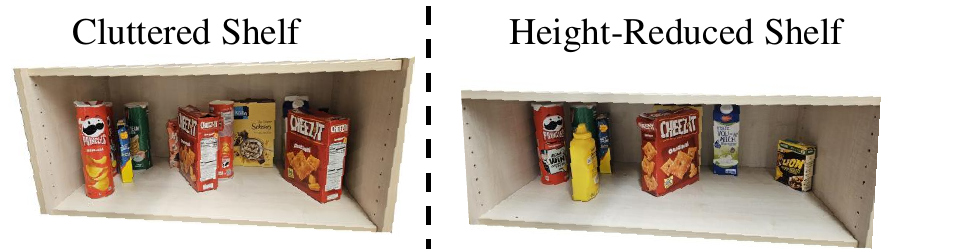}
\caption{Illustration of our two evaluation scenes:  a cluttered environment with a partially filled shelf providing sufficient vertical clearance, and a height-reduced environment where the clearance is reduced to 26.5\,cm, increasing collision risk.
}
  \label{fig:envs}
  \vspace{-10px}
\end{figure}
\subsection{Unified Pick-and-Place Reasoning on Real Hardware}
To evaluate the effectiveness of our placeability-informed unified pick-and-place pipeline (\textbf{UniP}), we conduct quantitative real-world experiments on a UR5e platform and compare against two sequential baselines and one ablated variant.
Each pipeline was tested on five objects, performing three pick-and-place executions per object in two shelf configurations~(see Fig.~\ref{fig:envs}).
The first scenario (Cluttered Shelf Environment) consists of a partially filled shelf \mbox{(76 × 38 × 33\,cm)} containing 10 objects and allowing placement of all target objects without severe spatial constraints while still introducing realistic clutter. 
In the second scenario (Height-Reduced Environment), the available vertical clearance is reduced to 26.5\,cm, significantly increasing the risk of collisions and requiring more precise grasp and placement selection.
Qualitative results are shown in the supplemental material.
\subsubsection{\textbf{Performance Comparison with Baselines}}
The first baseline \textbf{Grasp-RP} employs a sequential pick-then-place strategy, i.e., selects the highest-scoring grasp predicted by GPD~\cite{GPD_TenPas_2017} and then assigns a random placement pose from $\mathcal{T}_P$~(see Sec.~\ref{sec:placement_sampling}), while preserving the object’s original orientation.
The second baseline \textbf{Grasp-MO} follows the same sequential structure but allows six object orientations, while omitting both stability and placement-conditioned graspability. 
This baseline isolates the contribution of multi-orientation reasoning without explicit feasibility evaluation.
Finally, we evaluate \textbf{UniP-NoStab}, an ablated variant of our pipeline that follows the unified pick-and-place formulation but excludes the probabilistic stability term, allowing us to quantify the impact of stability reasoning within the unified pipeline.
We evaluate each method according to success-rate and failure types. The results are shown in Table~\ref{tab:multi_env_eval_extended}.

In the cluttered environment, our \textbf{UniP} achieves the highest success rate 93.4\% with only a single failure caused by grasp execution. 
In contrast, the sequential baseline \mbox{\textbf{Grasp-RP}} reaches only 46.6\%, with most failures occurring after successful grasping, where no feasible placement pose was found due to environmental constraints caused by the grasp direction. 
This demonstrates that grasp quality alone is insufficient when placement feasibility is not considered.
Allowing multiple placement orientations of the object in \textbf{Grasp-MO} improves performance to 60.0\%, confirming that orientation flexibility can increase placement opportunities, but frequent failures due to no possible place-grasps and unstable configurations remain.
Importantly, even if we include stability reasoning the failure case of no placement feasibility remains.
The unified ablation \mbox{\textbf{UniP-NoStab}} achieves 86.8\%, showing that jointly reasoning about grasp and placement feasibility already provides a substantial improvement. However, unstable placements still occur, highlighting the importance of modeling stability.
The advantage of our full method becomes especially present in the height-reduced environment. Here, \textbf{UniP} maintains a high success rate of \textbf{86.8\%}, while sequential baselines degrade significantly. 
\mbox{\textbf{Grasp-RP}} drops to 26.6\%, and \mbox{\textbf{Grasp-MO}} fails almost entirely, as most grasps that successfully lifted the object could not be placed due to collisions or insufficient clearance.
Even the unified ablation \textbf{UniP-NoStab} decreases to 60.0\%, with failures primarily caused by unstable placements.

Overall, these results confirm that the components of our method are necessary, i.e., \textit{stability reasoning prevents physically invalid placements, while unified grasp–placement feasibility reduces execution failures. }
Together, they enable consistently high performance across diverse objects even in constrained cluttered environments.
\begin{figure}[t]
\centering
\setlength{\tabcolsep}{2pt}
\setlength{\fboxsep}{0pt}

\resizebox{0.98\linewidth}{!}{%
\begin{minipage}{\linewidth}
\centering

\begin{minipage}[t]{0.49\linewidth}
    \centering
    \fbox{\includegraphics[width=\linewidth, trim={0 20 0 0}, clip]{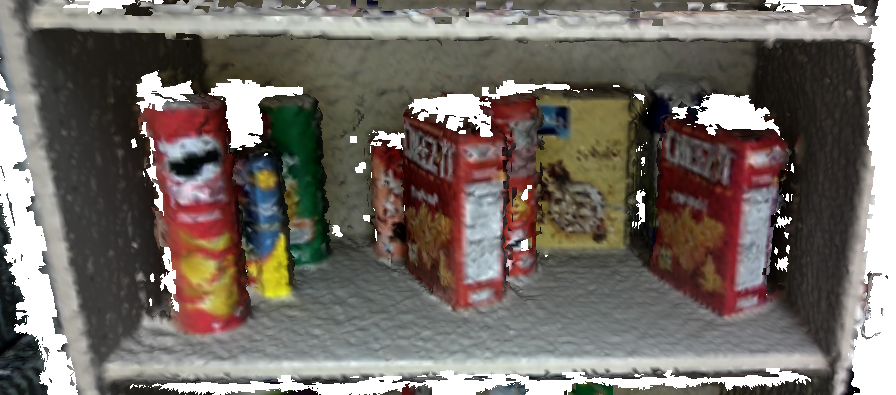}}
\end{minipage}\hfill
\begin{minipage}[t]{0.49\linewidth}
    \centering
    \includegraphics[width=0.8\linewidth, trim={0 20 0 0}, clip]{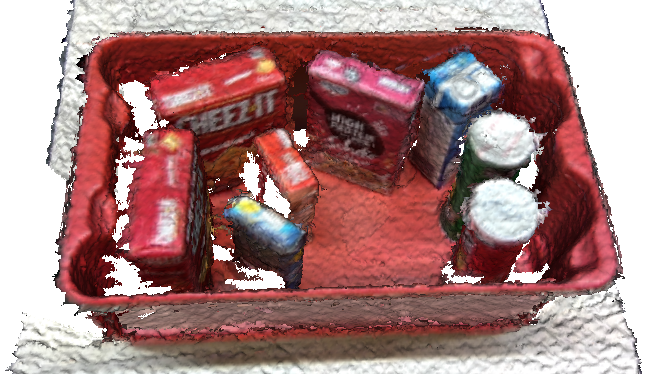}
\end{minipage}

\vspace{3pt}

\newcommand{\gimgw}{0.235\linewidth}

\begin{tabular}{cccc}
    \fbox{\includegraphics[width=\gimgw]{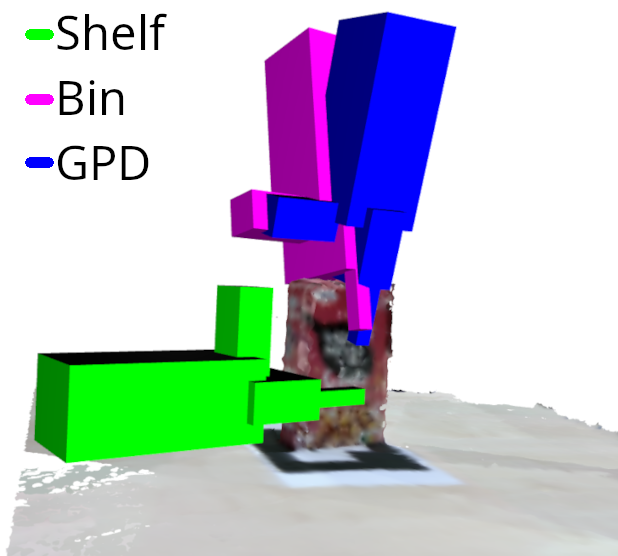}}
    & \fbox{\includegraphics[width=\gimgw]{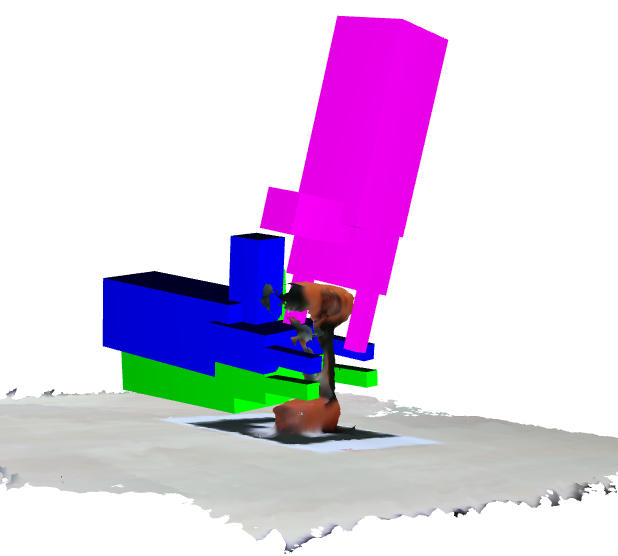}}
    & \fbox{\includegraphics[width=\gimgw]{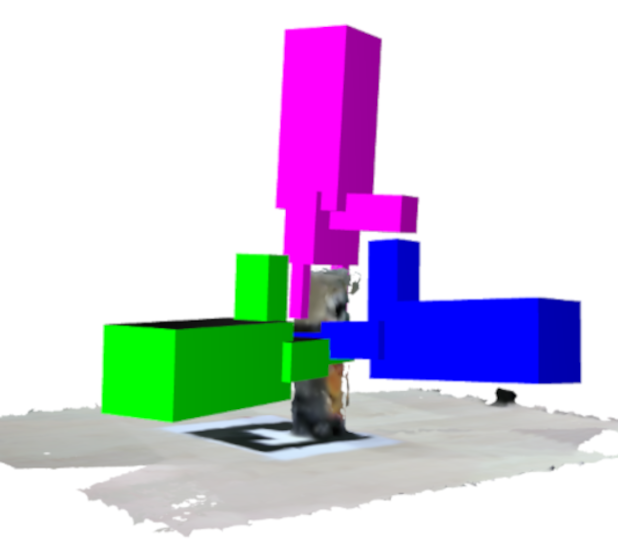}}
    & \fbox{\includegraphics[width=\gimgw]{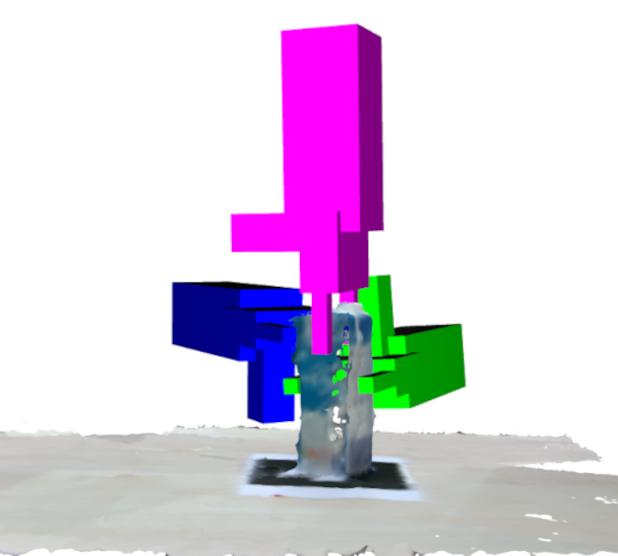}}\\
    Cereals & Drill & Pringles & Milk
    
\end{tabular}

\end{minipage}%
}
\vspace{-5px}
\caption{
Qualitative grasp comparison for two placement scenes: a shelf (top left) and a bin (top right).
Grasp prediction candidates are shown for four perceived objects. 
Blue indicates the original GPD ranking~\cite{GPD_TenPas_2017}, while green (shelf) and magenta (bin) indicate the grasps re-ranked using our placeability metric. 
By accounting for the target placement during grasp selection, our method prioritizes grasps that remain executable (i.e., collision-free with the shelf or bin structure) and physically stable at the desired location over grasp-optimal but placement-infeasible configurations. 
}
\label{fig:placeability_shelf_bin}
\vspace{-8px}
\end{figure}
\subsubsection{\textbf{Runtime Analysis}}
To assess the online performance of our full pick-and-place reasoning pipeline, we measured the total execution time across all five objects and all trials in the cluttered shelf environment, resulting in a full pick-and-place runtime of $182.47 \pm 21.43$~s. 
The majority of this time ($169.17 \pm 21.34$~s) is spent in task-external execution overhead, dominated by robot motion planning, motion execution, and  grasp computation that are unrelated to our placeability computation. In contrast, the perception and reasoning modules add only a small fraction of the total runtime.
Our TSDF-based reconstruction pipeline requires $8.85 \pm 0.63$~s, while the sampling of placement candidate poses takes only  $0.70 \pm 0.02$~s. 
Finally evaluating our placeability metric, including probabilistic stability and placement feasibility, takes $4.98 \pm 0.33$~s, indicating that the proposed metric can be computed online with negligible impact on overall task time, as multiple parts can be computed in parallel, while providing the stability and feasibility reasoning needed for reliable manipulation in cluttered scenes.

\subsubsection{\textbf{Place-Grasp Evaluation}}
To evaluate the adaptability of our unified reasoning approach, we assess how grasp selection changes when explicitly conditioned on different target placement regions, i.e., a shelf and a bin. 
Therefore, we run the full pipeline on four objects in those two scenarios.
The results, shown in Fig.~\ref{fig:placeability_shelf_bin}, demonstrate that our placeability-based rescoring systematically adapts grasp selection to the placement context without requiring any parameter changes. 
In constrained environments, grasps with greater clearance and improved placement feasibility are prioritized. 
In contrast, grasp-only ranking via GPD frequently leads to grasps, which can complicate motion planning and increase the risk of collision during placement.
This behavior indicates that the proposed metric adapts grasp selection when necessary while preserving high-quality grasp choices in simple placement settings.

\section{Conclusion}
\label{sec:conclusion}
In this work, we introduced a probabilistic placeability metric that enables model-free unified pick-and-place reasoning directly from noisy partial point clouds while identifying physically feasible object poses across diverse orientations and support conditions. 
By jointly evaluating probabilistic stability, and placement-conditioned graspability, the proposed formulation allows robots to select grasp–placement pairs that are both executable and physically stable without relying on CAD models or predefined placement poses.
Experiments on YCB objects demonstrated that our stability formulation accurately predicts tipping behavior beyond planar assumptions, capturing edge-proximity and inclined-surface stability directly from partial observations. 
Compared to UOP-Net~\cite{noh2024learning}, our method achieved higher placement accuracy, i.e. lower rotational and translational deviations after placement.
Finally, real-robot experiments in two changeling scenarios confirm that
our full unified reasoning pipeline outperforms sequential baselines and an ablation without stability reasoning in terms of success rate, confirming that both unified grasp–placement feasibility and probabilistic stability estimation are essential for robust execution.

\printbibliography

\end{document}